\documentclass[10pt,twocolumn,letterpaper]{article}

\usepackage{wacv}
\usepackage{times}
\usepackage{epsfig}
\usepackage{graphicx}
\usepackage{amsmath}
\usepackage{amssymb}

\usepackage[nolist,nohyperlinks]{acronym}
\usepackage{booktabs}
\usepackage{subcaption}
\usepackage{pifont}
\usepackage{multirow}
\usepackage{makecell}

\def\wacvPaperID{173} 

\wacvfinalcopy 

\ifwacvfinal
\def\assignedStartPage{9876} 
\fi

\ifwacvfinal
\usepackage[breaklinks=true,bookmarks=false]{hyperref}
\else
\usepackage[pagebackref=true,breaklinks=true,colorlinks,bookmarks=false]{hyperref}
\fi

\ifwacvfinal
\else
\fi

\acrodef{dl}[DL]{deep learning}
\acrodef{cnns}[CNNs]{convolutional neural networks}
\acrodef{ntk}[NTK]{neural tangent kernel}
\acrodef{cntk}[CNTK]{convolutional neural tangent kernel}

\begin{document}

\title{On the Effectiveness of Neural Ensembles for Image Classification with Small Datasets}

\author{Lorenzo Brigato\\
Sapienza University of Rome\\
{\tt\small brigato@diag.uniroma1.it}
\and
Luca Iocchi\\
Sapienza University of Rome\\
{\tt\small iocchi@diag.uniroma1.it}
}

\maketitle

\begin{abstract}
Deep neural networks represent the gold standard for image classification.
However, they usually need large amounts of data to reach superior performance.
In this work, we focus on image classification problems with a few labeled examples per class and improve data efficiency by using an ensemble of relatively small networks. \\
For the first time, our work broadly studies the existing concept of neural ensembling in domains with small data, through extensive validation using popular datasets and architectures.
We compare ensembles of networks to their deeper or wider single competitors given a total fixed computational budget.
We show that ensembling relatively shallow networks is a simple yet effective technique that is generally better than current state-of-the-art approaches for learning from small datasets.
Finally, we present our interpretation according to which neural ensembles are more sample efficient because they learn simpler functions.
\end{abstract}

\section{Introduction}
\label{sec:intro}

The advent of \ac{dl} has revolutionized the computer vision field \cite{DBLP:journals/nature/LeCunBH15}. However, the cost to reach high recognition performances involves the collection and labeling of large quantities of images.
This requirement can not always be fulfilled since it may happen that collecting images is extremely expensive or not possible at all.
Different approaches have been proposed by the research community to mitigate the necessity of training data, tackling the problem from different perspectives.
Among them of particular interest are definitely transfer and few-shot learning \cite{bengio2012deep}, \cite{tan2018survey} \cite{DBLP:journals/corr/abs-1810-03548}.
Still, these approaches rely on a large set of image-annotation pairs on which re-usable representations can be learnt.

In this work, we propose the use of neural ensembles composed of smaller networks to tackle the problem of learning from a small sample and show the superiority of such methodology.
Similarly to what has been done in recent works, we benchmark the approaches by varying the number of data points in the training sample while keeping it low with respect to the current standards of computer vision datasets \cite{DBLP:conf/iclr/AroraD0SWY20}, \cite{barz2020deep}.
Due to its great difficulty, this problem is still unsolved and hardly experimented
despite its primary importance.

It has been shown that large \ac{cnns} can handle overfitting and generalize well even if they are severely over-parametrized  \cite{DBLP:journals/corr/abs-1710-05468}, \cite{DBLP:journals/corr/abs-1805-12076}.
A recent study has also empirically shown that such behavior might also be valid in the case of tiny datasets, making large nets a viable choice even when the training sample is limited \cite{bornscheinsmall}.
On the other hand, a well-known technique to reduce model variance is to average predictions from a set of weak learners (e.g. random forests \cite{breiman2001random}).
An ensemble of low-bias decorrelated learners, combined with randomized inputs and prediction averaging, generally mitigates overfitting.
Despite the high popularity of neural ensembles, our study is the first one that systematically trains them from scratch on datasets with few samples per class and gives empirical evidence of their advantage over state-of-the-art approaches.
Our strict experimental methodology compares ensembles with state-of-the-art methods and single networks keeping a fair comparison in terms of model resources.

Therefore, we study ensembles of \ac{cnns} in small-data tasks by a) fixing a computational budget and b) comparing them to corresponding deeper or wider single variants.
According to our empirical study, ensembles are preferable over wider networks that are in turn better than deeper ones.
Moreover, obtained results make ensembles of smaller networks a strong baseline and an advantageous basic building block for future works that will tackle the problem of learning from small datasets.

In summary, the contributions of our work are the following: \emph{i}) we systematically study neural ensembles with small datasets and show that they generally outperform state-of-the-art methods; \emph{ii}) we make a structured study comparing ensembles of smaller-scale networks and their computationally equivalent single competitors with increased depth or width; \emph{iii}) we explain the better performance of ensembles by showing their bias towards learning less complex functions.

\section{Related Work}
\label{sec:related_work}


\paragraph{Learning from limited data} is still largely unsolved.
As previously said, few works have tried to tackle the problem of training \ac{dl} architectures with a small number of samples due to its difficulty.
A series of works focused on the classification of vector data and mainly used the UCI Machine Learning Repository as a benchmark \cite{fernandez2014we}, \cite{olson2018modern}, \cite{DBLP:conf/iclr/AroraD0SWY20}.

However, in this work, we are interested in image classifiers.
Transformations that synthesize new images are a straightforward approach to improve generalization (e.g. data augmentation \cite{shorten2019survey}).
Whit limited data, it was shown that even standard data augmentation is fundamental to raise the performance of neural networks \cite{brigato2020close}.
More particular domains like face recognition instead, should rely on had-hoc approaches \cite{hu2017frankenstein}.
Training generative models (e.g. GANs) represents an attractive approach to increase the dataset size and consequently, performance \cite{liu2019generative}.
However, training a generative model might be computationally intensive or present severe challenges in the small sample domain (e.g. overfitting, mode collapse) \cite{karras2020training}. \cite{azurigenerative} proposed to generate images directly optimizing a fixed latent space to obviate such difficulties.
\cite{DBLP:conf/iccci/Rueda-PlataRG15} suggested training \ac{cnns} with a greedy layer-wise method, analogous to that used in unsupervised deep networks.
\cite{bietti2019kernel} presented diverse kernel-based regularization techniques to regularize deep networks.
\cite{barz2020deep} proposed the use of the cosine loss to prevent overfitting. 
On the other hand, \cite{DBLP:conf/iclr/AroraD0SWY20} performed experiments with \ac{cntk}.
Finally, \cite{xu2020towards} proposed a rotation-invariant and scale-invariant architecture to improve generalization and data efficiency.

\paragraph{Ensembles of neural networks} are widely successful for improving both the accuracy and predictive uncertainty of very different predictive tasks, ranging from supervised learning \cite{hansen1990neural} to few-shot learning \cite{DBLP:conf/iccv/DvornikMS19}.\\
Recent studies empirically show that deep ensembles can outperform single models with equivalent computational budgets on image classification tasks \cite{dutt2018coupled}, \cite{chirkova2020deep}, \cite{geiger2020scaling}, \cite{wasay2021more}. 
We have in common with these works the outcome, i.e. the superiority of neural ensembles over single networks.
However, our study spans different directions of the problem space.
Indeed, \cite{geiger2020scaling} suggested the possibility of obtaining better performance with ensembles but without performing an extensive empirical evaluation with state-of-the-art architectures.
\cite{dutt2018coupled} mainly focused their attention on the possible strategies to combine the predictions of the ensemble members. 
Differently from \cite{chirkova2020deep}, we compare neural ensembles with single networks not only of the same depth but also of the same width and greater depth.
Furthermore, given few training examples as in our study, we empirically show that claims regarding the advantage of depth over width do not hold \cite{wasay2021more}.
Indeed, we present evidence that deeper networks tend to struggle in the case of small datasets.
Finally, in contrast with previous work, we provide a more solid interpretation of the better performance of ensembles.
We base our reasoning on a sensitivity metric that was proposed to measure neural function complexity \cite{novak2018sensitivity}. \\
A drawback of neural ensembles regards the computational cost that generally grows linearly with the ensemble dimension.
Some works mitigated such problem in the field of continual and supervised learning \cite{DBLP:conf/iclr/WenTB20}, \cite{DBLP:conf/mlsys/WasayHLCI20}.
We share this concern by fixing a budget, making the computational cost of training ensembles equivalent or lower to the one of a single network \cite{wasay2021more}.

\section{Problem Definition and Methodology}


\subsection{Image recognition with small data}

We are dealing with a problem in which we assume to have a dataset of samples and corresponding labels \(\mathcal{D} = \{(\mathbf{x}_{1}, \mathbf{y}_{1}) \dots, (\mathbf{x}_{s}, \mathbf{y}_{s})\}\) where each label \(\mathbf{y}\) is a one-hot vector that is encoding \(K\) different classes.\\
Furthermore, we are assuming that \(\mathcal{D}\) is balanced and that the number of samples per class is equal to \(N\) and relatively small.
The notion of "small" is intrinsically related to the difficulty of the task.
Indeed, it may be possible that good recognition performance can be reached with few samples per class.
Yet, we quantify this lack of data by assuming to have at most tens/hundreds of samples per class.
Concerning the dimensions of current computer vision benchmarks, this can be generally described as a reduction of one/two orders of magnitudes.\\
In general, \(\mathbf{x}\) might be a vector of any size. 
In our work, we are tackling a classification problem where each input sample is an image, hence, \(\mathbf{x} \in \mathbb{R}^{H \times W \times D}\).  
The objective is to learn the input-output mapping \(\mathbf{y} = f_{\theta}(\mathbf{x})\) with \(f(\cdot)\) representing a function approximator and \(\mathbf{\theta}\) its parameters.

\subsection{Neural ensembles}

The operational pipeline of neural ensembles consists of training multiple networks individually and then averaging their predictions obtaining, in the end, better results than single estimates.
More precisely, since we are implementing a weighted-average ensemble, we define a set \(\mathcal{M} = \{g_{\theta_{m}}(\cdot): m = 1, \dots M\}\) of \(M\) functions.
In our particular implementation, the ensemble is homogeneous, therefore all functions \(g\) are equal.
At testing time, given an input sample \(\mathbf{x}\), the ensemble prediction is going to be the mean of the predictions coming from each member in \(\mathcal{M}\) scaled by a function \(\phi\) depending on the type of the loss used. More precisely, \(\mathbf{y} = f_{\theta}(\mathbf{x}) = \frac{1}{M} \sum_{m=1}^{M} \phi \left( g_{\theta_{m}}(\mathbf{x})\right)\).
The weights \(\theta_{m}\) are randomly initialized and ensemble members update their parameters receiving the same input batch.
Throughout the optimization process, the members are independent, therefore each network will minimize its own loss  \(\mathcal{L}(\mathbf{y}, \phi (g_{\theta_{m}}(\mathbf{x})))\).

\subsection{Networks' design space}

In our analysis, we will evaluate neural ensembles and compare them with deep or wide single networks.
Given an ensemble \(\mathcal{M}\) with \(M\) base members \(g_{\theta_{m}}\), each network has depth \(d\) and width \(w\).
The set of networks account for a total computational budget \(B\).
To perform a fair ranking, the competing single networks have the same \(B\).
More in detail, deep or wide single competitors have, respectively, the same \(w\) but increased depth \(d' > d\) or the same \(d\) but increased width \(w' > w\).
We express the budget as the number of floating-point operations (FLOPs) to make a prediction.
This is a standard metric that measures algorithmic/model complexity.\footnote{In our implementation, we used Tensorflow APIs to compute the FLOPs of networks.}
Note that two networks of the same family and layout not only have a similar number of FLOPs but also of trainable parameters.

\section{Experimental Evaluation}
\label{sec:exp}


\subsection{Evaluation metrics}
\label{sec:perf_metrics}

In our work, we will use two metrics to evaluate the performance of the trained networks.\\
The first one is the classic accuracy over the testing set.
Indeed, we will use benchmarks that are balanced in terms of samples per class, making accuracy an appropriate metric to measure generalization.
To ensure consistency of the results, we perform five independent runs and then compute mean and standard deviations.\\
The second metric measures the sensitivity of the network to input perturbations and is computed as the Frobenius norm of the input-output Jacobian (\(||J(\mathbf{x})||_{F}\)) given \(J(\mathbf{x}) = \partial{f_{\theta}}(\mathbf{x})/\partial\mathbf{x}^{T}\).
We will use this metric to support our interpretation of the effectiveness of neural ensembles with limited data.
\cite{sokolic2017robust} theoretically showed the connection between input-change robustness and generalization, and \cite{novak2018sensitivity} provided empirical evidence of this result.
On a qualitative level, this metric tracks the
first-order term of the Taylor expansion of the function around a point.
Therefore, low sensitivity values (i.e. low norm of input-output Jacobian) suggest smoother functions less influenced by input perturbations.

\subsection{Datasets}
\label{sec:ds}
We perform our study on five popular benchmarks for
image classification: CIFAR-10, CIFAR-100 \cite{krizhevsky2009learning}, SVHN \cite{netzer2011reading}, Stanford Dogs \cite{khosla2011novel} and CUB-200 \cite{wah2011caltech}. More details regarding the datasets are available in the Appendix.

In this work, we are interested in benchmarking the capabilities of neural ensembles with limited data.
To this end, we sub-sample the training sets of the chosen datasets to match our specifications.
More in detail, for all the datasets, except CUB-200, the number of samples per class is varied in the set \(\{10, 50, 100, 250\}\).
For Stanford Dogs, we stop at 100 samples having reached the original size of the training dataset.
On the other hand, since CUB-200 has at most 30 samples per class, we train with \(\{5, 10, 20, 30\}\) samples per class.
A very similar procedure was already proposed in \cite{barz2020deep}, and \cite{azurigenerative}.\\
Note that the test datasets are not changed and remain fixed throughout all the evaluations.

\subsection{Experimented networks}
\label{sec:compared_models}

\begin{table*}[t]
    \centering
    \caption{Results of neural ensembles and state-of-the-art approaches evaluated on sub-sampled versions of popular datasets. 
    We report in \(\text{\color{blue}{blue}}\) our ensembles and in black the other methods.
    We highlight the highest value in black bold font.
    Values correspond to \(mean \pm std\).}
    \label{tab:sota}
    
    \subcaption{Comparison with the best of Table 4 from \cite{bietti2019kernel} (\(\dagger\)).}
    \resizebox{\textwidth}{!}
	{\scriptsize\begin{tabular}{cccc|cccc}
\toprule
\textbf{Dataset} &\textbf{Aug.}                         & \textbf{Model} & \textbf{N} = 100 
& \textbf{Dataset} & \textbf{Aug.}                          & \textbf{Model} & \textbf{N} = 500        \\
\midrule
\multirow{8}{*}{\rotatebox[origin=c]{90}{{CIFAR-10}}} & \multirow{2}{*}{-}  & VGG-11 + grad-\(l_{2}\) + SN proj\(^{\dagger}\)  & 46.88       &\multirow{8}{*}{\rotatebox[origin=c]{90}{{CIFAR-10}}}   & \multirow{2}{*}{-}          & VGG-11 + PGD-\(l_{2}\) + SN proj\(^{\dagger}\)       & 64.50          \\

&  & \color{blue}{10 VGG-5}   & \textbf{52.16 \(\pm\) 0.38} & &  &  \color{blue}{10 VGG-5} & \textbf{68.59 \(\pm\) 0.21} \\

\cmidrule{4-4} \cmidrule{8-8}

& \multirow{2}{*}{-} &  ResNet-18 + \(||\nabla f||^{2}\)\(^{\dagger}\) & 44.97   & & \multirow{2}{*}{-} & ResNet-18 + \(||f||^{2}_{\delta}\) SN proj\(^{\dagger}\)  & 59.03          \\
&  &  \color{blue}{20 ResNet-8} & \textbf{52.66 \(\pm\) 0.45}          & &  &  \color{blue}{20 ResNet-8} & \textbf{69.49 \(\pm\) 0.36} \\
\cmidrule{4-4} \cmidrule{8-8}

& \multirow{2}{*}{+}   & VGG-11 + grad-\(l_{2}\) + SN proj\(^{\dagger}\) & 55.32         &          & \multirow{2}{*}{+} & VGG-11 + grad-\(l_{2}\)\(^{\dagger}\) & 75.38          \\

&                        & \color{blue}{10 VGG-5} & \textbf{57.73 \(\pm\) 0.52}     &     &      & \color{blue}{10 VGG-5} & \textbf{76.26 \(\pm\) 0.09} \\

\cmidrule{4-4} \cmidrule{8-8}

& \multirow{2}{*}{+}         & ResNet-18 + grad-\(l_{2}\)\(^{\dagger}\) & 49.30                   & & \multirow{2}{*}{+}     & ResNet-18 grad-\(l_{2}\) + SN proj\(^{\dagger}\) & 77.73          \\
&  & \color{blue}{20 ResNet-8} & \textbf{63.64 \(\pm\)0.61}          &  &   & \color{blue}{20 ResNet-8} & \textbf{81.78 \(\pm\) 0.19} \\
\bottomrule
\end{tabular}

}
	
	\vskip 0.1in
	\subcaption{Comparison with the best of Table 2 from \cite{DBLP:conf/iclr/AroraD0SWY20} (\(\sharp\))}
    \resizebox{\textwidth}{!}
	{\LARGE\begin{tabular}{ccccccccccc}
\toprule
\textbf{Dataset} &\textbf{Aug.} &\textbf{Model} & \textbf{N} = 1 & \textbf{N} = 2  & \textbf{N} = 4 & \textbf{N} = 8 & \textbf{N} = 16   & \textbf{N} = 32 & \textbf{N} = 64 & \textbf{N} = 128\\

\midrule

\multirow{3}{*}{\rotatebox[origin=c]{90}{{\large{CIFAR-10}}}} & \multirow{3}{*}{-} & ResNet-34\(^{\sharp}\) & 14.59 \(\pm\) 1.99 &	17.5 \(\pm\) 2.47&	19.52 \(\pm\) 1.39&	23.32 \(\pm\) 1.61&	28.3 \(\pm\) 1.38	&33.15 \(\pm\) 1.2&	41.66 \(\pm\) 1.09&	49.14 \(\pm\) 1.31\\
& & CNTK\(^{\sharp}\) & 15.33 \(\pm\) 2.43&	18.79 \(\pm\) 2.13&	21.34 \(\pm\) 1.91&	25.48 \(\pm\) 1.91&	30.48 \(\pm\) 1.17&	36.57 \(\pm\) 0.88&	42.63 \(\pm\) 0.68&	48.86 \(\pm\) 0.68\\
& & \color{blue}{20 ResNet-8} & \textbf{16.73 \(\pm\) 1.02} &	\textbf{20.66 \(\pm\) 0.53} &	\textbf{24.47 \(\pm\) 0.9} &	\textbf{29.54 \(\pm\) 0.91} &	\textbf{34.55 \(\pm\) 0.92} &	\textbf{40.48 \(\pm\) 1.28} &	\textbf{47.5 \(\pm\) 0.24} &	\textbf{55.04 \(\pm\) 0.46}\\
\bottomrule
\end{tabular}}
    
    \vskip 0.1in
	\subcaption{Comparison with results from Figure 6 and Table 1 of \cite{azurigenerative} (\(\ast\)), Figure 10 of \cite{xu2020towards} (\(\diamond\)) and Figure 2 of \cite{barz2020deep} (\(\ddagger\)). The values of the plots were provided by the authors or extracted with an on-line tool.}
    \resizebox{\textwidth}{!}
	{\scriptsize\begin{tabular}{cccccccc}
\toprule
\textbf{Dataset} &\textbf{Aug.} &\textbf{Model} & \textbf{N} = 10 & \textbf{N} = 50  & \textbf{N} = 100 & \textbf{N} = 250\\

\midrule

\multirow{3}{*}{\rotatebox[origin=c]{90}{{\tiny{CIFAR-10}}}} & + \& Cutout & 1 WRN-28-10\(^{\ast}\) & 27.41 \(\pm\)  0.57	&  46.39 \(\pm\)  0.39	&  57.97 \(\pm\)  0.31	& 73.86 \(\pm\)  0.68\\
& GLICO & 1 WRN-28-10\(^{\ast}\) & 32.12 \(\pm\) 0.38	& 54.32 \(\pm\) 1.35	& 67.19 \(\pm\) 0.36	&  \textbf{78.23} \\
& +++ & \color{blue}{20 WRN-10-5} & \textbf{34.01 \(\pm\) 0.27}	& \textbf{59.24 \(\pm\) 0.23}	& \textbf{67.96 \(\pm\) 0.48}	& 78.15 \(\pm\) 0.26 \\
\midrule

\multirow{6}{*}{\rotatebox[origin=c]{90}{{\scriptsize{CIFAR-100}}}} & GLICO & 1 WRN-28-10\(^{\ast}\) & \textbf{28.55 \(\pm\) 0.4}	& \textbf{52.95 \(\pm\) 0.2}	& \textbf{64.27 \(\pm\)0.04} & -- \\
& +++ & \color{blue}{20 WRN-10-5} & 24.47 \(\pm\) 0.48	& 51.09 \(\pm\) 0.45 &	61.82 \(\pm\) 0.31 & --\\
\cmidrule{4-8}

& \multirow{1}{*}{+} & Data-Eff. CNN\(^{\diamond}\) & 14 & 36 &	52.67 & -- &\\
& \multirow{2}{*}{+ \& Cutout} & 1 ResNet-110-32 (cosine)\(^{\ddagger}\) & 18.47 & 42.86 &	55.89 & 68.06 &\\

&  & 1 ResNet-110-32 (cosine + xe)\(^{\ddagger}\) & 16.99 & 	45.44 & 	57.14 & 	68.66 &\\

& +++ & \color{blue}{20 ResNet-8-32 (xe)} & \textbf{23.74 \(\pm\) 0.69} & \textbf{51.14 \(\pm\) 0.35}	& \textbf{61.41 \(\pm\) 0.36}	& \textbf{70.68 \(\pm\) 0.12}\\

\midrule

& & & \textbf{N} = 5 & \textbf{N} = 10  & \textbf{N} = 20 & \textbf{N} = 30\\
\midrule
\multirow{4}{*}{\rotatebox[origin=c]{90}{{\scriptsize{CUB-200}}}} & \multirow{2}{*}{+ \& Cutout} & 1 ResNet-50 (cosine)\(^{\ddagger}\) & 10.65 & 	23.20 & 	53.14 & 	67.6 & \\

&  & 1 ResNet-50 (cosine + xe)\(^{\ddagger}\) & 11.25 & 	30.82 & 	55.18 & 	\textbf{67.95} & \\

& \multirow{2}{*}{+} & \color{blue}{5 ResNet-10 (cosine)} & 15.77 \(\pm\) 0.05	& 31.9 \(\pm\) 0.08	& 55.97 \(\pm\) 0.19	& 66.69 \(\pm\) 0.05\\

&  & \color{blue}{5 ResNet-10 (cosine + xe)} & \textbf{16.3 \(\pm\) 0.09}	& \textbf{32.94 \(\pm\) 0.01}	& \textbf{56.03 \(\pm\) 0.51}	& 66.87 \(\pm\) 0.01\\


\bottomrule
\end{tabular}}
    
\end{table*}

We considered four different families of convolutional architectures: VGG \cite{DBLP:journals/corr/SimonyanZ14a}, ResNet \cite{he2016deep}, Wide ResNet (WRN) \cite{DBLP:conf/bmvc/ZagoruykoK16} and DenseNet \cite{huang2017densely}.
We tested the first three on the CIFAR datasets, while DensNet on SVHN and Stanford Dogs.
We also tested the Imagenet version of ResNet on CUB-200.

For ResNet and VGG on CIFAR, the maximum computational budget was set to be roughly close to the FLOPs of a single ResNet-110 and VGG-9.
On CUB-200, due to the larger image size, the budget was equivalent to the one of a ResNet-50 with 64 base filters.
For WRN we considered a single WRN-28-10. 
For DenseNet on SVHN, we have chosen the complexity of DenseNet-BC-52 while, for Stanford Dogs, DenseNet-BC-121.\\
On the CIFAR datasets with ResNets, we also set two smaller computational budgets matching the FLOPs of a single ResNet-26 and ResNet-50 with 16 base filters.\\
As base members for the ensembles, we chose to have ResNet-8 with 16 base filters, WRN-10 with width factor equal to 5, WRN-10 with width factor equal to 5, ResNet-10 with 64 base filters, VGG-5 with 32 base filters, DenseNet-BC-16 with growth rate (\(k\)) equal to 12 and DenseNet-BC-62 with \(k\) = 56.\\
For more details regarding the computational complexity and structure of the networks refer to the Appendix.

\subsection{Training set-up}

To train our networks we used standard data pre-processing and optimization techniques. 
We set three different augmentation levels based on conventional transformations to ensure that results were not biased in this respect.
We added to the standard random crop and flipping policy widely used (+), random color distortion (++), and finally random erasing (+++).\\

More details regarding training schedules and augmentation strategies are available in the Appendix.

\section{Results}
\label{sec:results}

\subsection{Comparing with the state of the art}

\begin{table*}[t]
    \centering
    \caption{Comparison between deep/wide \ac{cnns} and ensembles of less complex networks.
    \(M\) indicates the number of networks in the ensemble and \(N\) the number of training samples per class.
    These results are obtained using medium data augmentation (++). Means and standard deviations are obtained from five independent runs.}
    \vskip 0.1in
    \label{tab:all_ds}
    \resizebox{\textwidth}{!}
    {\scriptsize\begin{tabular}{cccccccc}
\toprule
\textbf{Dataset}& \textbf{Budget} \scriptsize{(MFLOPs)} &\textbf{Model} & \textbf{M} &\textbf{N} = 10 & \textbf{N} = 50  & \textbf{N} = 100 & \textbf{N} = 250 \\
\midrule

\multirow{6}{*}{\rotatebox[origin=c]{90}{{CIFAR-10}}}&\multirow{3}{*}{\rotatebox[origin=c]{0}{{\(\sim\) 9.5}}}& ResNet-110-16 & 1 & 28.65 \(\pm\) 1.06	& 43.68 \(\pm\) 1.29	& 51.16 \(\pm\) 0.96	& 63.0 \(\pm\) 1.7 \\

& & ResNet-8-72 & 1 & 32.12 \(\pm\) 0.28	& 54.58 \(\pm\) 0.36	& 62.82 \(\pm\) 0.62	& 74.76 \(\pm\) 0.1\\

& & ResNet-8-16 & 20 & \textbf{34.9 \(\pm\) 1.13}	& \textbf{58.11 \(\pm\) 0.42}	& \textbf{66.95 \(\pm\) 0.58}	& \textbf{76.89 \(\pm\) 0.11}\\

\cmidrule{5-8}

&\multirow{3}{*}{\rotatebox[origin=c]{0}{{\(\sim\) 4.5}}} & VGG-9-32	& 1 & 28.5 \(\pm\) 1.6	& 42.25 \(\pm\) 1.29	& 47.51 \(\pm\) 1.68	& 58.07 \(\pm\) 0.44\\

& & VGG-5-76	& 1 & 32.99 \(\pm\) 1.57	& 48.06 \(\pm\) 0.12	& 55.41 \(\pm\) 0.73	& 65.45 \(\pm\) 0.24\\

& & VGG-5-32	& 5 & \textbf{34.15 \(\pm\) 1.54}	& \textbf{51.5 \(\pm\) 0.41}	& \textbf{60.28 \(\pm\) 0.63}	& \textbf{70.73 \(\pm\) 0.33}\\

\midrule

\multirow{6}{*}{\rotatebox[origin=c]{90}{{CIFAR-100}}}&\multirow{3}{*}{\rotatebox[origin=c]{0}{{\(\sim\) 9.5}}}& ResNet-110-16 & 1 & 13.13 \(\pm\) 0.54	&29.53 \(\pm\) 0.26	&38.23 \(\pm\) 1.61	&53.17 \(\pm\) 0.73 \\
& & ResNet-8-72 & 1 & 19.65 \(\pm\) 1.09	&45.16 \(\pm\) 0.02	&54.85 \(\pm\) 0.19	&\textbf{65.47 \(\pm\) 0.44}\\
& & ResNet-8-16 & 20 & \textbf{22.62 \(\pm\) 0.82}	&\textbf{48.76 \(\pm\) 0.35}	& \textbf{57.75 \(\pm\) 0.11}	&\textbf{65.41 \(\pm\) 0.15}\\

\cmidrule{5-8}

&\multirow{3}{*}{\rotatebox[origin=c]{0}{{\(\sim\) 4.5}}} & VGG-9-32	& 1 & 10.36 \(\pm\) 0.53	&22.95 \(\pm\) 0.79	&30.97 \(\pm\) 0.64	&42.92 \(\pm\) 0.51\\
& & VGG-5-76	& 1 & 14.35 \(\pm\) 0.3	&28.25 \(\pm\) 0.09	&36.84 \(\pm\) 0.27	&45.49 \(\pm\) 0.31\\
& & VGG-5-32	& 5 & \textbf{17.47 \(\pm\) 0.38}	&\textbf{36.39 \(\pm\) 0.13}	&\textbf{46.05 \(\pm\) 0.22}	&\textbf{55.21 \(\pm\) 0.31}\\

\midrule
\multirow{3}{*}{\rotatebox[origin=c]{90}{SVHN}} & \multirow{3}{*}{\rotatebox[origin=c]{0}{{\(\sim\) 0.5}}}  & DenseNet-BC-52, k=12 & 1 & \textbf{16.72 \(\pm\) 1.75}	& 78.42 \(\pm\) 1.19	& 86.52 \(\pm\) 0.24	& 89.6 \(\pm\) 0.7\\
& & DenseNet-BC-16, k=30 & 1 & 16.44 \(\pm\) 3.8 &	76.41 \(\pm\) 1.65 &	85.41 \(\pm\) 0.52 &	89.28 \(\pm\) 0.06 \\
& & DenseNet-BC-16, k=12 & 6 & 14.01 \(\pm\) 2.5 &	\textbf{82.02 \(\pm\) 1.67} &	\textbf{87.73 \(\pm\) 0.44} &	\textbf{91.61 \(\pm\) 0.32} \\

\midrule

\multirow{3}{*}{\rotatebox[origin=c]{90}{St. Dogs}} &  \multirow{3}{*}{\rotatebox[origin=c]{0}{{\(\sim\) 14.3}}} & DenseNet-BC-121, k=32 & 1 & 6.93 \(\pm\) 0.86	& 28.32 \(\pm\) 1.33	& 47.7 \(\pm\) 1.17 & --\\
& & DenseNet-BC-62, k=56 & 1 & 7.33 \(\pm\) 0.35 &	29.25 \(\pm\) 0.76 &	47.82 \(\pm\) 0.83 & --\\
& & DenseNet-BC-62, k=32 & 3 & \textbf{8.42 \(\pm\) 0.02} &	\textbf{35.12 \(\pm\) 0.68} &	\textbf{53.39 \(\pm\) 0.45} & --\\

\bottomrule
 
\end{tabular}}
\end{table*}

First, we show the effectiveness of deep ensembles comparing them with state-of-the-art techniques that were proposed for learning from a small sample. 
More precisely, diverse kernel-based regularizations proposed in \cite{bietti2019kernel}, the CNTK tested in \cite{DBLP:conf/iclr/AroraD0SWY20}, classifiers augmented with the generative model GLICO in \cite{azurigenerative}, the cosine loss of \cite{barz2020deep} and Data-Efficient \ac{cnns} \cite{xu2020towards}.
The first two works have chosen to test their models on sub-sampled versions of CIFAR-10 while the second two used CIFAR and CUB-200 datasets.
The results are shown in Table \ref{tab:sota}.

\cite{bietti2019kernel} evaluate their models with and without standard augmentation composed of random cropping and mirroring.
One hundred and five hundred samples per class comprised their training sets.
For sake of completeness, we also evaluate our ensembles in the latter case, despite the training set already contains a consistent number of samples.
They trained a VGG-11 and ResNet-18 with various regularization techniques.
To keep a fair comparison, we train an ensemble of 10 VGG-5 and 20 ResNet-8 and compare them with the corresponding regularized architectures tested in \cite{bietti2019kernel}.
Note, however, that with the default widths, both VGG-11 and ResNet-18 have more training parameters than our ensembles.
Moreover,  in the original work, authors tuned the hyper-parameters of their methods on a held-out validation set.\footnote{This is a simplifying assumption that inevitably increases training set dimension.}
We note from Table \ref{tab:sota} that our ensembles always outperform the respective models from a minimum of \(\sim 2\%\) to a maximum of \(\sim 14\%\) average accuracy. 

\cite{DBLP:conf/iclr/AroraD0SWY20} used a very restrictive training protocol varying the number of samples per class from 1 to 128 and not using any kind of data augmentation.
Here, we compare our ensemble of 20 ResNet-8 with their baseline ResNet-34 and proposed CNTK.
Our ensemble has fewer total trainable parameters than a single ResNet-34.
Also, in this case, gains in terms of average accuracy are noticeable and in the range of approximately \(1\% - 7\%\).

\begin{figure*}[t]
  \centering
  \includegraphics[scale=0.35]{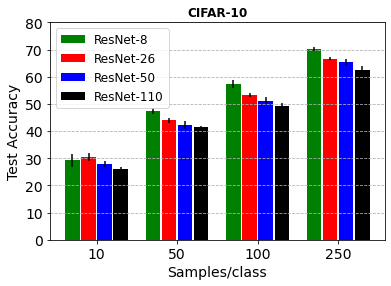}
  \hspace{0.01cm}
  \includegraphics[scale=0.35]{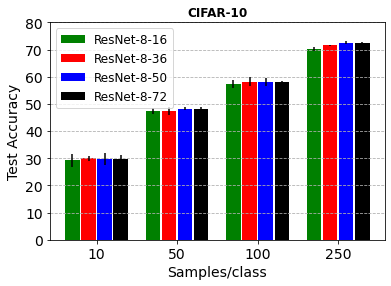}
    \hspace{0.01cm}
  \includegraphics[scale=0.35]{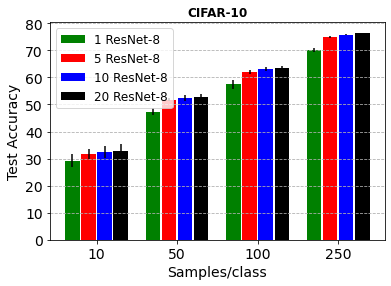}
  \vspace{0.1cm}
  \includegraphics[scale=0.35]{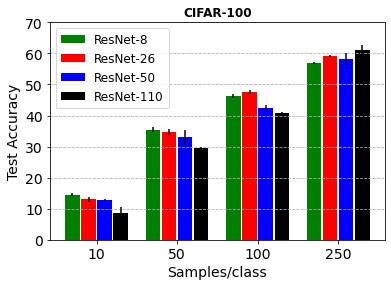}
  \hspace{0.15cm}
  \includegraphics[scale=0.35]{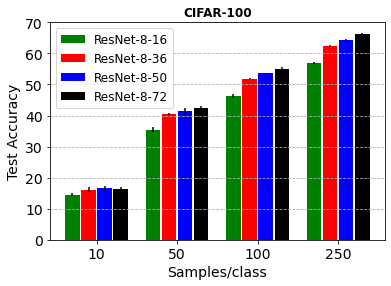}
    \hspace{0.15cm}
  \includegraphics[scale=0.35]{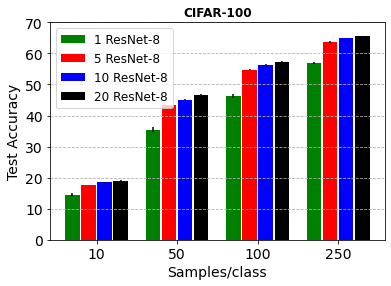}
  
  \caption{Results considering, respectively, the variation of depth/width in single networks, and the number of networks in the ensemble for ResNets on CIFAR datasets trained with standard augmentation. \textbf{Left}: allocation of resources in network's depth. \textbf{Middle}: allocation of resources in network's width. \textbf{Right}: allocation of resources in multiple networks.}
  \label{img:dep_wid_ens}
\end{figure*}

\cite{azurigenerative} tested WRN-28-10 on CIFAR datasets and ResNet-50 on CUB-200 augmented by GLICO.
We do not compare our approach with GLICO on the latter dataset since, in that case, they used a pre-trained ResNet-50 classifier.
Note also that GLICO uses a pre-trained-VGG based perceptual loss to synthesize images.
Therefore, differently from our method, GLICO requires, in any case, a pre-trained model.
We see from Table \ref{tab:sota} that, on CIFAR-10, ensembles are comparable to or better than the single network augmented with GLICO (gains of \(\sim 2\% - 5\%\) average accuracy).
This fact is surprising considering the previously cited advantages that GLICO has over our standard ensembles.
On the other hand, on the larger CIFAR-100, the single WRN plus GLICO outmatches 20 WRNs by roughly \(2\% - 5\%\) average accuracy.

\cite{barz2020deep} trained a ResNet-110 with 32 base filters on CIFAR-100, ResNet-50 on CUB-200 with multiple losses, and one-hot or semantic labels using standard augmentation plus Cutout.
In our case, we did not train ensembles with any kind of a priori knowledge, therefore, we considered their results obtained with the standard one-hot labels.
A combination of the cosine and cross-entropy loss scored the best values in their experiments.
We compared an ensemble of 20 ResNet-8-32 trained with the cross-entropy loss to their single ResNet-110-32 on CIFAR-100 and show that we get improvements of roughly \(2\% - 6\%\) average accuracy.
On CUB-200 we train an ensemble of 5 ResNet-10 to match their ResNet-50.
As in \cite{barz2020deep} we obtained poor results with the cross-entropy on this fine-grained dataset.
Therefore, we perform the comparison using the cosine and smoothed cosine losses.
In this case, as well, we get comparable results using the full dataset and gains of up to \( \sim 5\%\) average accuracy on smaller versions.

Finally, we report the results of Data-Efficient \ac{cnns} obtained on CIFAR-100 in \cite{xu2020towards}.
Our ensemble of 20 ResNet-8-32 has comparable complexity to this competitor in terms of parameters but better data efficiency. Indeed, gains in terms of average accuracy reach \(15\%\).

In fair comparisons, neural ensembles outperform the competing techniques.
In the more complex challenge with a generative model, ensembles can still be a valid choice.
Indeed, on the one hand, they require less computational needs.
On the other hand, they could be still strengthened by the generative model.
These results clearly show that standard ensembles with equivalent model complexity is a simple yet effective way to reduce model variance with few training samples.

\subsection{Comparing deep or wide single nets with ensembles}

In this section, we report the results of the ensembles and analyze their relation with their deeper or wider single competitors.
From Table \ref{tab:all_ds}, we note that ensembles are the best option in terms of testing accuracy for almost all cases.

We notice that deeper networks struggle with small datasets.
On CIFAR benchmarks, ResNet-8 and VGG-5 ensembles obtain very large gains over deeper architectures ranging from a minimum of \(\sim 3\%\) to a maximum of \(\sim 15\%\) average accuracy.
To be sure that these results are not just due to a poor regularization of the deepest model, we also made tests with more aggressive data augmentation which are available in the Appendix.
However, results remained consistent with the ones with medium augmentation.
Considering DenseNets, we appreciate a similar trend, with more moderate but still significant gains, from \(\sim 1.5\%\) to \(\sim 6\%\) average accuracy.

On the other hand, wider networks seem to handle the lack of training data better than deeper models.
Still, model averaging and independent training has a clear advantage over wider single networks.
ResNet, VGG and DenseNet ensembles generalize better than wider nets in terms of average accuracy from a minimum of \(\sim 1\%\) to a maximum of \(\sim 5\%\).
Additional experiments and discussions regarding depth, width, and ensembles will follow in the next sections.

\subsection{Comparing design spaces}

\begin{figure*}[t]
  \centering
  \includegraphics[scale=0.31]{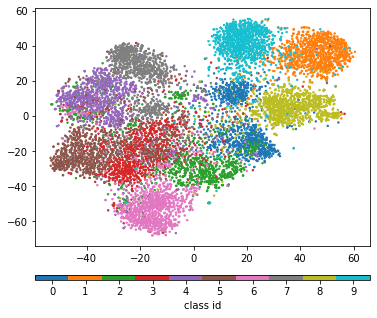}
  \hspace{0.001cm}
  \includegraphics[scale=0.31]{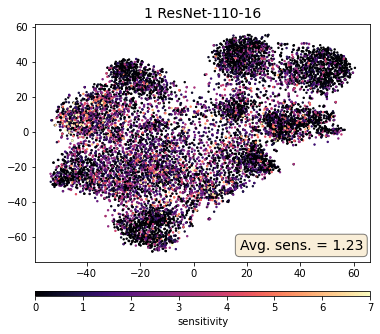}
    \hspace{0.01cm}
  \includegraphics[scale=0.31]{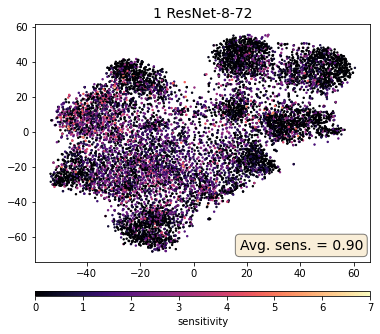}
      \hspace{0.01cm}
  \includegraphics[scale=0.31]{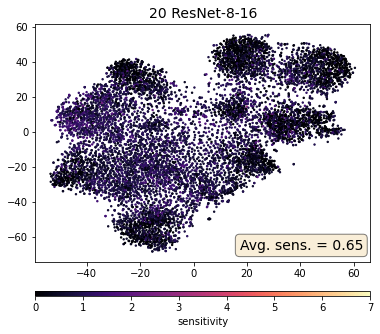}
  \vspace{0.1cm}
  \includegraphics[scale=0.31]{images_response/class_ids.png}
  \hspace{0.01cm}
  \includegraphics[scale=0.31]{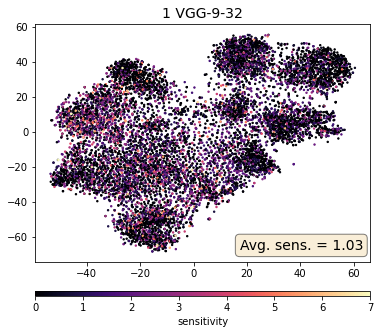}
    \hspace{0.01cm}
  \includegraphics[scale=0.31]{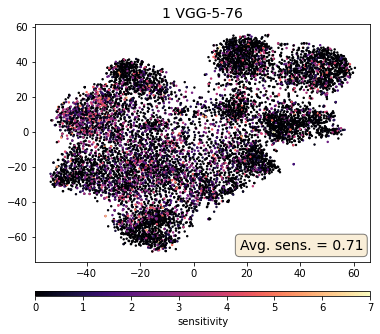}
    \hspace{0.01cm}
  \includegraphics[scale=0.31]{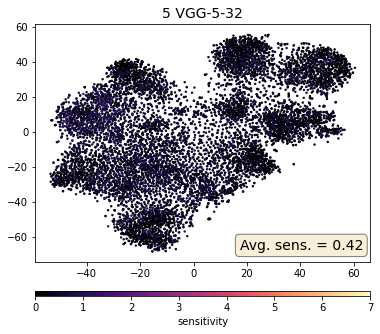}
   \vspace{0.1cm}
  \includegraphics[scale=0.31]{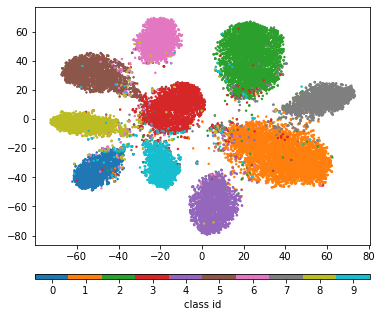}
  \hspace{0.01cm}
  \includegraphics[scale=0.31]{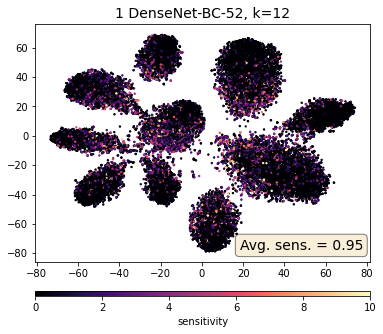}
    \hspace{0.01cm}
  \includegraphics[scale=0.31]{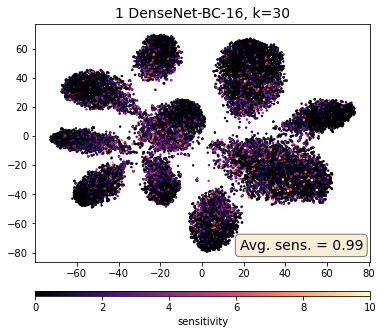}
  \hspace{0.01cm}
  \includegraphics[scale=0.3]{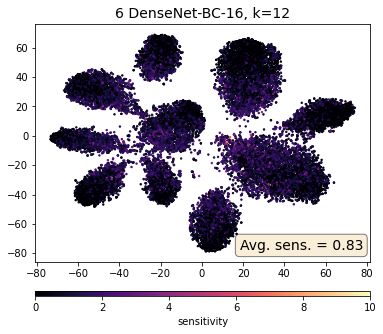}
  \caption{tSNE visualization of networks sensitivity to testing samples fixing the feature space of an \textit{oracle} network. The first two rows and the third one respectively correspond to CIFAR-10/SVHN with 100/50 samples per class.}
  \label{img:sens_dep_wid_ens}
\end{figure*}

We believe it is worth discussing in more detail the experiments with ResNets on the CIFAR datasets to better understand what is the best design choice among depth or width in single networks or the number of members in ensembles.
To provide a complete picture, we set, as mentioned in Section \ref{sec:compared_models}, additional computational budgets in terms of FLOPs that are lower than the FLOPs of a single ResNet-110.
The results of this analysis are visible in Figure \ref{img:dep_wid_ens} but also present in table format in the Appendix.
In the plots, the four budgets are represented by the different colors assuming green as the lowest budget (single ResNet-8-16) and black as the highest budget (e.g. single ResNet-110-16).
Therefore, an increment of testing accuracy from green to black indicates a favorable design direction.

A first result is that increasing the computational budget in the depth direction is generally not beneficial (left plots in Figure \ref{img:dep_wid_ens}).
Indeed, we note that the deepest model is always the worst, except for the case of CIFAR-100 with more samples.
This is reasonable since it was empirically shown that deep networks coupled with larger datasets can learn very complex functions that translate in good representational capacity \cite{DBLP:conf/nips/MontufarPCB14}.\\
Investing on network's width is slightly better (middle plots in Figure \ref{img:dep_wid_ens}).
On CIFAR-100, wider networks outperform the baseline model.
This result is expected.
Indeed, on a more complex problem, more capacity gives an advantage to wide networks.
On the other hand, on CIFAR-10, wide models perform comparably well to ResNet-8-16.

Finally, we note that ensembles have a clear advantage over the single ResNet-8-16 on both CIFAR-10 and CIFAR-100, with larger gains obtained on the latter (right plots in Figure \ref{img:dep_wid_ens}).
This makes ensembles the most preferable approach having the possibility of increasing the computational budget.
We also note that the largest gap is generally between the single network and 5 ResNet-8 proving that the performance gain is already considerable with relatively small ensembles.

\subsection{Interpreting data efficiency of neural ensembles}
\label{sec:explain}

In this final section, we provide our interpretation of why ensembles perform better than competitors with small datasets.\\
Given our design choices, deep or wide single nets and ensembles of networks potentially have the same complexity expressed in terms of free parameters.
However, it was shown that this general conformity does not automatically ensure equivalently performing models \cite{rasmussen2001occam}.
Also in our case, different designs of the same architecture led to different performances.
Therefore, we are led to believe that the functions implemented by ensembles are superior in terms of generalization and should differ from the ones learned by deeper or wider single nets.
A straightforward way to characterize the learned input-output mapping is to measure its complexity.
We anticipated in Section \ref{sec:perf_metrics} that the complexity of the learned function on the data manifold can be qualitatively measured by the Frobenius norm of the input-output Jacobian.
Low values of \(||J(\mathbf{x})||_{F}\) indicate smooth output variations given perturbations of \(\mathbf{x}\), i.e. low local sensitivity of the learnt function.
A more global estimate can be obtained computing the mean sensitivity on the test sample \cite{novak2018sensitivity}.
To better understand this phenomenon and visually show it, we perform the following analysis.

First, we train \textit{oracle} networks on full datasets to fix a perceptually robust representation space.
In this manner, perceptually similar inputs are generally mapped close to each other in the \textit{oracle} feature space.
Secondly, we feed the test samples to the \textit{oracle} and extract features right before the classification layer.
Then, we perform a 2D tSNE to decrease their dimensions.
Now, each test sample corresponds to a 2D vector \((p_{1}, p_{2})\).
In the left plots of Figure \ref{img:sens_dep_wid_ens}, we assign a different color to each of them depending on the class to which they belong.
In other words, each point is a 3D tuple \((p_{1}, p_{2}, c)\) with \(c\) being the class color.
As a third step, we consider networks trained on smaller datasets and compute \(||J(\mathbf{x})||_{F}\) for each testing sample obtaining the local sensitivity at \(\mathbf{x}\).
To visualize it on the available version of the input space, i.e. the test set, we assign \(||J(\mathbf{x})||_{F}\) to previously fixed positions \((p_{1}, p_{2})\) obtaining a scatter plot of a 3D tuple \((p_{1}, p_{2}, ||J(\mathbf{x})||_{F})\) being \(||J(\mathbf{x})||_{F}\) the color (right plots of Figure \ref{img:sens_dep_wid_ens}).
Note also that running tSNE for each network, obtaining independent embeddings (i.e. different \((p_{1}, p_{2})\)), would not have been possible for ensembles since their members do not share a common representation space.

In Figure \ref{img:sens_dep_wid_ens} we show the just introduced visualizations along with the average sensitivity over testing sets of ResNets/VGGs trained with 100 samples per class on CIFAR-10, and DenseNets trained with 50 per class on SVHN.
We also report additional results concerning network sensitivity in the Appendix.
Ensembles have the lowest sensitivity even in regions where there is the highest perceptual similarity among samples of different classes.
We believe that this quality is a direct consequence of model averaging and independent training.
Qualitatively speaking, as we change the input, the predictions of the overall ensemble model are smoothed thanks to the intrinsic property of an ensemble to have members with decorrelated errors.
Indeed, even if a subset of networks undergoes large input-output oscillations, there are still the remaining ones that can balance out their influence.

We notice that deeper ResNet-110 and VGG-9 have higher sensitivity than their wider competitors.
While single DenseNets are closer in terms of average sensitivity (\(0.95\) and \(0.99\)).
Previous works studied the expressive power of ReLU networks and argued that the complexity of the computed function grows exponentially with the depth and polynomially with the width \cite{pascanu2013number}, \cite{DBLP:conf/nips/MontufarPCB14}, \cite{raghu2017expressive}.
Our empirical results confirm the bias of deeper nets to learn more complex functions.
While this might be an advantage on larger and more complex datasets, it does not seem so with scarce data.
Reasonably, at test time networks experience a larger input variance with respect to scenarios with large datasets.
These input perturbations grow exponentially in the depth of the network \cite{raghu2017expressive} causing poorer performance.

On the other hand, our shallower and wider tested networks seem to learn less sensible functions. \cite{nguyen2018optimization} analyzed the influence of depth and width on \ac{cnns} and predicted that network width smoothes the optimization landscape of the loss.
We align with this claim since also after increasing the width on very small training sets (e.g. CIFAR-10 with 50 samples per class), networks perform comparably well or better than thinner ones (Figure \ref{img:dep_wid_ens}).\\

To sum up, taking previous arguments and results into account, it is reasonable to believe that deeper networks are more prone to overfit on small datasets due to their inherent property of learning more complex functions that scale exponentially with depth.
Wide networks seem to be more suited to learn less complex functions and to improve the optimization landscape.
Finally, ensembles add the advantage of model averaging and independent training to the low sensitivity of a single shallow network.
Note that ensemble members need to have enough representational capacity to fit the available training sample.
In the case of limited data, this can also be reached with relatively small computational budgets as we have seen in our experiments.
Yet, a smaller ensemble of larger-size nets or a single wider model might outperform a larger ensemble of smaller-size nets as the task complexity increases \cite{wasay2021more}.

\section{Conclusions}

In this paper, we have investigated the capability of deep ensembles on image classification problems with limited data.
Through comparing ensembles given a fixed computational budget, our results show that ensembles are generally better than current state-of-the-art approaches and their deeper or wider single competitors.

Furthermore, we analyzed the function space of tested models using the popular sensitivity metric based on the input-output Jacobian.
Functions implemented by ensembles reported lower sensitivity to input perturbations.
Therefore, we conjecture that neural ensembles are more suitable for small datasets because they learn simpler functions.

The encouraging results presented in this work suggest the potential of using ensembles for small datasets and promise further developments of this line of research in the field of small data.

{\small
\bibliographystyle{ieee_fullname}
\bibliography{egbib}
}

\end{document}


\title{Appendix}
\appendix

\maketitle

\section{Additional details on the datasets}
\label{sec:appendix_dataset}

\paragraph{CIFAR-10} is an established computer vision benchmark consisting of color images coming from \(10\) different classes of objects and animals \cite{krizhevsky2009learning}.
The dataset has originally \(50,000\) training images and \(10,000\) testing images with both sets balanced.
The input images are of size \(32 \times 32 \times 3\). 

\paragraph{CIFAR-100} is the more complex version of CIFAR-10 being composed of 100 classes and containing the same number of training and testing samples \cite{krizhevsky2009learning}.
The training/testing splits and input dimensionalities are equal to the ones of CIFAR-10.

\paragraph{SVHN} is a real-world image dataset semantically similar to MNIST since contains images of digits \cite{netzer2011reading}.
The popular cropped-version sets originally have $73,257$ training and $26,032$ testing images of dimension $32 \times 32 \times 3$.

\paragraph{Stanford Dogs} is a fine-grained categorization problem consisting of 120 species of dogs \cite{khosla2011novel}. The dataset has been built using samples and annotation from ImageNet and contains images with sides not smaller than 200 pixels.
The original training set contains \(100\) samples per class while the test set has in total \(8,580\) images.

\paragraph{CUB-200} is a fine-grained dataset contatining pictures of variable size of 200 bird species (mostly North American) \cite{wah2011caltech}.
The original training set is made of 5,994 images (almost 30 images per class) and the testing set of 5,794.

\section{Additional details on the architectures}
\label{sec:comp_complexity}

In this section, we provide more specifications about the used architectures and their computational complexities.

We used ResNet and DenseNet architectures proposed in the original works \cite{he2016deep} and \cite{huang2017densely}.
For the case of datasets with images of small size (e.g. CIFAR-10) the default base width of ResNets is equal to 16 and the growth rate of DenseNets is set to 12.
On the other hand, on larger input sizes, we adopted the architectures that were proposed in the original papers for Imagenet.
Therefore, ResNets have 64 base filters while DenseNets growth rate equal to 32.
In the case of DenseNets, to make smaller competitors and build an ensemble, we halved the number of dense blocks in each stack of DenseNet-BC-121 making a DenseNet-BC-52.

Finally, we removed the two densely connected layers at the bottom of VGG-11 to make a less computationally expensive VGG-9. Moreover, we set the base width to 32 filters instead of 64 to further decrease the computational load.
The shortest version, that we called VGG-5, is obtained by keeping a single convolutional layer for the first four original blocks.
Before the last dense layers, all VGG architectures are equipped with a dropout layer with drop-rate equal to 0.4.

In the case of WRNs (\cite{DBLP:conf/bmvc/ZagoruykoK16}), we indicated the width factor and used dropout with drop-rate equal to 0.3.   

In Table \ref{tab:flops} we report the computational complexity of single networks used in the corresponding dataset.
To obtain the computational complexity of ensembles, directly multiply the FLOPs of a single network by the number of total networks in the ensemble.

\begin{table}[H]
    \centering
    {\small\begin{tabular}{ccccc}
\toprule
Model & CIFAR-10 & CIFAR-100 & SVHN & Stanford Dogs\\
\midrule
ResNet-8-16 & 0.46 & 0.47 & -- & -- \\
ResNet-8-36 & 2.33 & 2.35 & -- & -- \\
ResNet-8-50 & 4.48 & 4.52 & -- & -- \\
ResNet-8-72 & 9.29 & 9.34 & -- & -- \\
ResNet-20-16 & 1.62 & 1.63 & -- & -- \\
ResNet-26-16 & 2.21 & 2.22 & -- & -- \\
ResNet-50-16 & 4.54 & 4.55 & -- & -- \\
ResNet-110-16 & 10.36 & 10.37 & -- & -- \\
\midrule
VGG-5-32 & 0.79 & 0.98 & -- & -- \\
VGG-5-76 & 4.41 & 4.85 & -- & -- \\
VGG-9-32 & 4.61 & 4.66 & -- & -- \\
\midrule
DenseNet-BC-16, k=12 & -- & -- & 0.091 & -- \\
DenseNet-BC-16, k=30 & -- & -- & 0.54 & -- \\
DenseNet-BC-52, k=12 & -- & -- & 0.54 & -- \\
\midrule
DenseNet-BC-62, k=32 & -- & -- & -- & 4.79 \\
DenseNet-BC-62, k=56 & -- & -- & -- & 14.36 \\
DenseNet-BC-121, k=32 & -- & -- & -- & 14.32 \\
\bottomrule

\end{tabular}}
    \caption{Computational complexity of single networks in terms of millions of FLOPS (MFLOPs).}
    \label{tab:flops}
\end{table}

\section{Additional details on training schedules}
\label{sec:exp_details}

In this section, we provide more details regarding the training set-up used to train our models.

\paragraph{Data pre-processing} In all experiments, input images are normalized subtracting the per-channel mean computed over the sub-sampled training set.

The augmentation strategies that we have used in this work are summarized in Table \ref{tab:augs}.
They correspond to widely used image transformations used in previous works \cite{he2016deep}, \cite{huang2017densely}.
For the case of small images, each image is padded 4 pixels on each side and a \(32 \times 32\) crop is randomly sampled from the padded image or its horizontal flip (+).

The addition of color distortion includes random hue, contrast, brightness and saturation (++).
For the highest level of data augmentation (+++), we further used pixel-level random erasing with the default parameters proposed in the original paper \cite{DBLP:conf/aaai/Zhong0KL020}.

On datasets with larger inputs (e.g. Stanford Dogs), all images were resized to \(256 \times 256\) pixels.
The random crop and flipping strategy generated patches of \(224 \times 224\) pixels that were then fed to the network.
At evaluation time, we applied a single central crop of size \(224 \times 224\).

\begin{table}[H]
    \centering
    \caption{Data augmentation strategies used in this work.}
    \label{tab:augs}
    \vskip 0.1in
    \scalebox{0.8}{\begin{tabular}{ccc}
\toprule
\textbf{Level} &\textbf{Symbol} & \textbf{Description} \\

\midrule
Absent  & -     & No augmentations \\
\cmidrule{3-3}
Standard     & +     & Random crop and horizontal flip \\
\cmidrule{3-3}
Medium  & ++    & \makecell{Random color distortion plus \\ all previous augmentations}\\
\cmidrule{3-3}
High    & +++   & \makecell{Random erasing plus \\ all previous augmentations} \\

\bottomrule
\end{tabular}}
\end{table}

\paragraph{Optimization} For training ResNets, WRNs and DenseNets, we use stochastic gradient descent (SGD) with weight decay and Nesterov momentum respectively set to \(10^{-4}\) and \(0.9\) as was done in the original papers \cite{he2016deep}, \cite{huang2017densely} and \cite{DBLP:conf/aaai/Zhong0KL020}.
We start with a learning rate of \(0.1\) and decrease it after 75\% of the total number of iterations by an order of magnitude.
We increased the number of iterations with the initial learning rate to be sure of decreasing it after having reached the training loss plateau. Furthermore, we have noticed that decreasing a second time the learning rate did not improve further the testing performance. 
For the deeper ResNet networks (i.e. ResNet-26, ResNet-50 and ResNet-110) we noticed some instability and larger variance in terms of testing accuracy on CIFAR-10.
For this reason, in that case, we trained them with the starting learning rate equal to \(0.01\) and then follow the same piece-wise schedule. We obtained more stable and faster convergence along with better results especially for the smaller datasets.
For the same reasons, we set the learning rate of ResNet-110 on CIFAR-100 to \(0.01\).
On the other hand, VGG architectures were trained with Adam optimizer and default hyper-parameters. All networks were fed with mini-batches of \(32\) images.

All ResNets, WRNs and DenseNets were trained for the same number of epochs. We noticed that VGG architectures needed less training epochs to converge (probably due to the faster convergence of Adam optimizer).
Assuming that the first element of the set corresponds to datasets with \(10\) samples per class and the last one to \(250\), architectures are trained, respectively for \(\{400, 300, 300, 250\}\) epochs.
On the other hand, for CUB-200, we trained for \(\{600, 400, 300, 300\}\) epochs considering the smaller number of images per class \(\{5, 10, 20, 30\}\).

\section{Additional experiments}
\label{sec:add_exp}

Finally, we provide additional experiments mentioned in the paper.

\paragraph{Table \ref{tab:dep-wid-ens_full}} we show additional tests that we run with different levels of data augmentation.
We note that networks generally obtain better performance as the level of data augmentation increases.
However, the advantage of ensembles over deeper networks is wide in any case. 
Further, ensembles still outperform wide networks by significant margins.
We note, that sometimes this gap is reduced by a couple of percentage points.
Thus, in some cases (e.g. ResNets on CIFAR-100) more aggressive augmentation has helped in a more significant way the widest model.
In this scenario, the wider single model that is more regularized can decrease the gap with or outperform the ensemble.
As we mentioned in the paper, this is happening if the available free parameters of ensembles members are not enough to fit the training set.

\paragraph{Table \ref{tab:ext_dep_wid_ens}} we show a different view of Figure 1 in the main paper to provide the comparison between ensembles and their single competitors given the same architecture at multiple computational budgets.
It can be seen that ResNet-8 ensembles outperform the competing wide ResNet-8 and deep ResNet in all cases except for CIFAR-100 and 250 samples per class (due to the previously mentioned reason).

\paragraph{Table \ref{tab:all_ds_sens}} we report additional results regarding the sensitivity analysis.
It can be seen that ensembles show the lowest sensitivity to input perturbations in almost all cases.

{\small
\bibliography{egbib}
\bibliographystyle{ieee_fullname}
}

\begin{table}[H]
    \centering
    \caption{Comparison in terms of test accuracy between deep/wide CNNs and ensembles of less complex networks considering different levels of data augmentation.
    \(M\) indicates the number of networks in the ensemble and \(N\) the number of training samples per class.
    Means and standard deviations are obtained from five independent runs.}
    \label{tab:dep-wid-ens_full}
    \vskip 0.05in
    \resizebox{\textwidth}{!}
    {\begin{tabular}{ccccccccc}
\toprule
\textbf{Dataset}& \textbf{Budget} \scriptsize{(MFLOPs)} & \textbf{Aug.} &\textbf{Model} & \textbf{M} &\textbf{N} = 10 & \textbf{N} = 50  & \textbf{N} = 100 & \textbf{N} = 250 \\
\midrule

\multirow{18}{*}{\rotatebox[origin=c]{90}{{CIFAR-10}}} & \multirow{3}{*}{\rotatebox[origin=c]{0}{{\(\sim\) 9.5}}}& \multirow{3}{*}{+}  & ResNet-110-16 & 1 & 26.06 \(\pm\) 0.56  &	41.32 \(\pm\) 0.58 &	49.21 \(\pm\) 1.04 &		62.5 \(\pm\) 1.49 \\
& & & ResNet-8-72 & 1 & 29.65 \(\pm\) 1.54 &	48.0 \(\pm\) 0.72 &	58.16 \(\pm\) 0.37 &	72.41 \(\pm\) 0.36 \\
& & & ResNet-8-16 & 20 & \textbf{32.83 \(\pm\) 2.39} &	\textbf{52.88 \(\pm\) 0.92} &	\textbf{63.64 \(\pm\) 0.61} &	\textbf{76.23 \(\pm\) 0.28} \\
\cmidrule{6-9}

&\multirow{3}{*}{\rotatebox[origin=c]{0}{{\(\sim\) 9.5}}} & \multirow{3}{*}{++} & ResNet-110-16	& 1 & 28.65 \(\pm\) 1.06	& 43.68 \(\pm\) 1.29	& 51.16 \(\pm\) 0.96	& 63.0 \(\pm\) 1.7\\
& & & ResNet-8-72	& 1 & 32.12 \(\pm\) 0.28	& 54.58 \(\pm\) 0.36	& 62.82 \(\pm\) 0.62	& 74.76 \(\pm\) 0.1\\
& & & ResNet-8-16	& 20 & \textbf{34.9 \(\pm\) 1.13}	& \textbf{58.11 \(\pm\) 0.42}	& \textbf{66.95 \(\pm\) 0.58}	& \textbf{76.89 \(\pm\) 0.11}\\
\cmidrule{6-9}

 & \multirow{3}{*}{\rotatebox[origin=c]{0}{{\(\sim 9.5\)}}}& \multirow{3}{*}{+++} & ResNet-110-16 & 1 & 29.05 \(\pm\) 0.8	& 45.77 \(\pm\) 0.64	& 57.81 \(\pm\) 0.96	& 67.47 \(\pm\) 1.21\\
& & & ResNet-8-72 & 1 & 32.29 \(\pm\) 1.47	& 55.54 \(\pm\) 0.94	& 65.28 \(\pm\) 1.21	& 75.61 \(\pm\) 0.62 \\
& & & ResNet-8-16 & 20 & \textbf{34.27 \(\pm\) 0.35}	& \textbf{58.15 \(\pm\) 0.57}	& \textbf{67.4 \(\pm\) 0.51} & 	\textbf{77.07 \(\pm\) 0.13} \\
\cmidrule{6-9}

& \multirow{3}{*}{\rotatebox[origin=c]{0}{{\(\sim\) 4.3}}} & \multirow{3}{*}{+} & VGG-9-32 & 1 &27.64 \(\pm\) 1.28	&41.74 \(\pm\) 0.11	&47.22 \(\pm\) 0.42	&56.36 \(\pm\) 1.52\\
& & & VGG-5-76 & 1 & 30.28 \(\pm\) 1.37	& 45.39 \(\pm\) 0.56	& 51.38 \(\pm\) 0.72	& 62.08 \(\pm\) 1.16\\
& & & VGG-5-32 & 5 & \textbf{31.69 \(\pm\) 1.03} &	\textbf{48.61 \(\pm\) 0.74} &	\textbf{57.18 \(\pm\) 0.61} &	\textbf{68.38 \(\pm\) 0.47}\\
\cmidrule{6-9}

& \multirow{3}{*}{\rotatebox[origin=c]{0}{{\(\sim\) 4.3}}} & \multirow{3}{*}{++} & VGG-9-32 & 1 &28.5 \(\pm\) 1.6	&42.25 \(\pm\) 1.29	&47.51 \(\pm\) 1.68	&58.07 \(\pm\) 0.44\\
& & & VGG-5-76 & 1 & 32.99 \(\pm\) 1.57	&48.06 \(\pm\) 0.12	&55.41 \(\pm\) 0.73	&65.45 \(\pm\) 0.24\\
& & & VGG-5-32 & 5 & \textbf{34.15 \(\pm\) 1.54}	&\textbf{51.5 \(\pm\) 0.41}	&\textbf{60.28 \(\pm\) 0.63}	&\textbf{70.73 \(\pm\) 0.33}\\
\cmidrule{6-9}

& \multirow{3}{*}{\rotatebox[origin=c]{0}{{\(\sim\) 4.3}}} & \multirow{3}{*}{+++} & VGG-9-32 & 1 &27.89 \(\pm\) 2.61	&43.14 \(\pm\) 0.77	&48.69 \(\pm\) 0.67	&59.17 \(\pm\) 0.72\\
& & & VGG-5-76 & 1 & 32.35 \(\pm\) 1.46	&48.54 \(\pm\) 0.83	&57.12 \(\pm\) 0.85	&65.62 \(\pm\) 0.81\\
& & & VGG-5-32 & 5 & \textbf{33.64 \(\pm\) 1.45}	&\textbf{52.37 \(\pm\) 0.11}	&\textbf{60.72 \(\pm\) 0.46}	&\textbf{70.95 \(\pm\) 0.59}\\

\midrule

\multirow{18}{*}{\rotatebox[origin=c]{90}{{CIFAR-100}}} & \multirow{3}{*}{\rotatebox[origin=c]{0}{{\(\sim\) 9.5}}} & \multirow{3}{*}{+}& ResNet-110-16 & 1 & 12.53 \(\pm\) 0.24	& 29.68 \(\pm\) 0.77	& 39.88 \(\pm\) 0.85	& 53.08 \(\pm\) 1.57\\
& & & ResNet-8-72 & 1   &16.51 \(\pm\)0.38	 &42.52 \(\pm\) 0.44	 &54.94 \(\pm\)0.8	 &\textbf{66.38 \(\pm\) 0.12} \\
& & & ResNet-8-16 & 20 & \textbf{18.92 \(\pm\) 0.38}	&\textbf{46.56 \(\pm\) 0.41}	&\textbf{57.37 \(\pm\) 0.05}	&65.56 \(\pm\) 0.21 \\
\cmidrule{6-9}

 & \multirow{3}{*}{\rotatebox[origin=c]{0}{{\(\sim\) 9.5}}} & \multirow{3}{*}{++}& ResNet-110-16 & 1 & 13.13 \(\pm\) 0.54	&29.53 \(\pm\) 0.26	&38.23 \(\pm\) 1.61	&53.17 \(\pm\) 0.73\\
& & & ResNet-8-72 & 1   &19.65 \(\pm\) 1.09	&45.16 \(\pm\) 0.02	&54.85 \(\pm\) 0.19	&\textbf{65.47 \(\pm\) 0.44}\\
& & & ResNet-8-16 & 20 & \textbf{22.62 \(\pm\) 0.82}	&\textbf{48.76 \(\pm\) 0.35}	&\textbf{57.75 \(\pm\) 0.11}	&65.41 \(\pm\) 0.15\\
\cmidrule{6-9}

 & \multirow{3}{*}{\rotatebox[origin=c]{0}{{\(\sim\) 9.5}}} & \multirow{3}{*}{+++}& ResNet-110-16 & 1 & 14.11 \(\pm\) 0.64	&32.04 \(\pm\) 0.71	&44.31 \(\pm\) 1.8	&58.62 \(\pm\) 0.57\\
& & & ResNet-8-72 & 1   &20.68 \(\pm\) 0.74	&46.15 \(\pm\) 0.07	&55.77 \(\pm\) 0.6	&\textbf{66.84 \(\pm\) 0.09}\\
& & & ResNet-8-16 & 20 & \textbf{22.04 \(\pm\) 1.05}	&\textbf{48.05 \(\pm\) 0.34}	&\textbf{56.97 \(\pm\) 0.25}	&63.59 \(\pm\) 0.18\\
\cmidrule{6-9}

& \multirow{3}{*}{\rotatebox[origin=c]{0}{{\(\sim\) 4.8}}}  & \multirow{3}{*}{+} & VGG-9-32 & 1 &10.22 \(\pm\) 0.38	& 23.94 \(\pm\) 0.34	& 31.04 \(\pm\) 0.59	& 42.09 \(\pm\) 1.01\\
& & & VGG-5-76 & 1 & 13.25 \(\pm\) 0.07	& 26.46 \(\pm\) 0.36	& 33.52 \(\pm\) 0.39	& 44.84 \(\pm\) 0.67\\
& & & VGG-5-32 & 5 & \textbf{16.29  \(\pm\) 0.57}	& \textbf{34.37 \(\pm\) 0.33}	& \textbf{44.04 \(\pm\) 0.17}	& \textbf{56.37 \(\pm\) 0.05}\\
\cmidrule{6-9}

& \multirow{3}{*}{\rotatebox[origin=c]{0}{{\(\sim\) 4.8}}}  & \multirow{3}{*}{++} & VGG-9-32 & 1 &10.36 \(\pm\) 0.53	&22.95 \(\pm\) 0.79	&30.97 \(\pm\) 0.64	&42.92 \(\pm\) 0.51\\
& & & VGG-5-76 & 1 & 14.35 \(\pm\) 0.3	&28.25 \(\pm\) 0.09	&36.84 \(\pm\) 0.27	&45.49 \(\pm\) 0.31\\
& & & VGG-5-32 & 5 & \textbf{17.47 \(\pm\) 0.38}	&\textbf{36.39 \(\pm\) 0.13}	&\textbf{46.05 \(\pm\) 0.22}	&\textbf{55.21 \(\pm\) 0.31}\\
\cmidrule{6-9}

& \multirow{3}{*}{\rotatebox[origin=c]{0}{{\(\sim\) 4.8}}}  & \multirow{3}{*}{+++} & VGG-9-32 & 1 &11.06 \(\pm\) 0.31	&23.78 \(\pm\) 0.47	&32.6 \(\pm\) 0.23	&43.44 \(\pm\) 0.04\\
& & & VGG-5-76 & 1 & 14.53 \(\pm\) 0.17	&29.34 \(\pm\) 0.23	&36.44 \(\pm\) 0.43	&44.77 \(\pm\) 0.52\\
& & & VGG-5-32 & 5 & \textbf{17.41 \(\pm\) 0.53}	&\textbf{36.26 \(\pm\) 0.3}	&\textbf{45.48 \(\pm\) 0.2}	&\textbf{53.21 \(\pm\) 0.21}\\
\midrule

\multirow{6}{*}{\rotatebox[origin=c]{90}{SVHN}} & \multirow{3}{*}{\rotatebox[origin=c]{0}{{\(\sim\) 0.5}}}  & \multirow{3}{*}{+}& DenseNet-BC-52, k=12 & 1 & \textbf{16.98 \(\pm\) 0.95}	&78.49 \(\pm\) 2.59	&86.67 \(\pm\) 0.21	&89.68 \(\pm\) 0.66\\
& & & DenseNet-BC-16, k=30 & 1 & 15.03 \(\pm\) 1.99	&65.99 \(\pm\) 0.79	&85.9 \(\pm\) 0.56	&88.97 \(\pm\) 0.33\\
& & & DenseNet-BC-16, k=12 & 6 & 14.51 \(\pm\) 1.61	&\textbf{80.25 \(\pm\) 0.94}	&\textbf{88.61 \(\pm\) 0.3}	&\textbf{91.46 \(\pm\) 0.17}\\
\cmidrule{6-9}

 & \multirow{3}{*}{\rotatebox[origin=c]{0}{{\(\sim\) 0.5}}}  & \multirow{3}{*}{++}& DenseNet-BC-52, k=12 & 1 & \textbf{16.72 \(\pm\) 1.75}	& 78.42 \(\pm\) 1.19	& 86.52 \(\pm\) 0.24	& 89.6 \(\pm\) 0.7\\
& & & DenseNet-BC-16, k=30 & 1 & 16.44 \(\pm\) 3.8 &	76.41 \(\pm\) 1.65 &	85.41 \(\pm\) 0.52 &	89.28 \(\pm\) 0.06 \\
& & & DenseNet-BC-16, k=12 & 6 & 14.01 \(\pm\) 2.5 &	\textbf{82.02 \(\pm\) 1.67} &	\textbf{87.73 \(\pm\) 0.44} &	\textbf{91.61 \(\pm\) 0.32} \\

\midrule

\multirow{6}{*}{\rotatebox[origin=c]{90}{St. Dogs}} &  \multirow{3}{*}{\rotatebox[origin=c]{0}{{\(\sim\) 14.3}}} & \multirow{3}{*}{+} & DenseNet-BC-121, k=32 & 1 &6.16 \(\pm\) 0.4	&30.07 \(\pm\) 0.17	&48.43 \(\pm\) 0.37&-- \\
& & & DenseNet-BC-62, k=56 & 1 &6.2 \(\pm\) 0.45	&28.94 \(\pm\) 0.66	&47.66 \(\pm\) 1.21& --\\
& & & DenseNet-BC-62, k=32 & 3 &\textbf{6.58 \(\pm\) 0.2}	&\textbf{33.33 \(\pm\) 1.05}	&\textbf{52.77 \(\pm\) 0.3}& --\\
\cmidrule{6-9}

&  \multirow{3}{*}{\rotatebox[origin=c]{0}{{\(\sim\) 14.3}}} & \multirow{3}{*}{++} & DenseNet-BC-121, k=32 & 1 & 6.93 \(\pm\) 0.86	& 28.32 \(\pm\) 1.33	& 47.7 \(\pm\) 1.17 & --\\
& & & DenseNet-BC-62, k=56 & 1 & 7.33 \(\pm\) 0.35 &	29.25 \(\pm\) 0.76 &	47.82 \(\pm\) 0.83 & --\\
& & & DenseNet-BC-62, k=32 & 3 & \textbf{8.42 \(\pm\) 0.02} &	\textbf{35.12 \(\pm\) 0.68} &	\textbf{53.39 \(\pm\) 0.45} & --\\

\bottomrule
\end{tabular}

}

\end{table}

\begin{table}[H]
    \centering
    \caption{This table provides a different view of Figure 1 in the main paper.
    In this format, the exact means and standard deviations are visible (test accuracy).
    The comparison is made at different computational budgets using the chosen ResNet architectures on CIFAR datasets with standard data augmentation (+).
    }
    \label{tab:ext_dep_wid_ens}
    \vskip 0.1in
    \resizebox{\textwidth}{!}{\scriptsize\begin{tabular}{cccccccc}
\toprule
\textbf{Dataset}& \textbf{Budget} \scriptsize{(MFLOPs)} &\textbf{Model} & \textbf{M} &\textbf{N} = 10 & \textbf{N} = 50  & \textbf{N} = 100 & \textbf{N} = 250 \\
\midrule

\multirow{10}{*}{\rotatebox[origin=c]{90}{{CIFAR-10}}} & 0.46 & ResNet-8-16	& 1 & 29.27 \(\pm\) 2.34&	47.37 \(\pm\) 1.26&	57.38 \(\pm\) 1.5&	70.12 \(\pm\) 0.84\\
\cmidrule{5-8}

&  \multirow{3}{*}{\rotatebox[origin=c]{0}{{\(\sim\) 2.3}}} & ResNet-26-16	& 1 & 30.49 \(\pm\) 1.51&	43.91 \(\pm\) 0.79&	53.29 \(\pm\) 0.84&	66.68 \(\pm\) 0.42\\
& & ResNet-8-36	& 1 & 29.94 \(\pm\) 0.87&	47.49 \(\pm\) 1.42&	58.24 \(\pm\) 1.49&	71.5 \(\pm\) 0.35\\
& & ResNet-8-16	& 5 & \textbf{31.75 \(\pm\) 1.78} &	\textbf{51.75 \(\pm\) 0.62} &	\textbf{61.98 \(\pm\) 0.81}&	\textbf{74.85 \(\pm\) 0.36}\\
\cmidrule{5-8}

&\multirow{3}{*}{\rotatebox[origin=c]{0}{{\(\sim\) 4.5}}} & ResNet-50-16	& 1 & 27.78 \(\pm\) 1.13 &	42.27 \(\pm\) 1.62	& 51.23 \(\pm\) 1.45	& 65.34 \(\pm\) 1.36\\
& & ResNet-8-50	& 1 & 29.66 \(\pm\) 2.32	& 48.13 \(\pm\) 0.55	& 58.1 \(\pm\) 1.6	& 72.54 \(\pm\) 0.67\\
& & ResNet-8-16	& 10 & \textbf{32.41 \(\pm\) 2.34}	& \textbf{52.44 \(\pm\) 1.02}	& \textbf{63.03 \(\pm\) 0.66}	& \textbf{75.84 \(\pm\) 0.31}\\
\cmidrule{5-8}

&\multirow{3}{*}{\rotatebox[origin=c]{0}{{\(\sim\) 9.5}}}& ResNet-110-16 & 1 & 26.06 \(\pm\) 0.56  &	41.32 \(\pm\) 0.58 &	49.21 \(\pm\) 1.04 &		62.5 \(\pm\) 1.49 \\
& & ResNet-8-72 & 1 & 29.65 \(\pm\) 1.54 &	48.0 \(\pm\) 0.72 &	58.16 \(\pm\) 0.37 &	72.41 \(\pm\) 0.36 \\
& & ResNet-8-16 & 20 & \textbf{32.83 \(\pm\) 2.39} &	\textbf{52.88 \(\pm\) 0.92} &	\textbf{63.64 \(\pm\) 0.61} &	\textbf{76.23 \(\pm\) 0.28} \\

\midrule

\multirow{10}{*}{\rotatebox[origin=c]{90}{{CIFAR-100}}} & 0.47 & ResNet-8-16	& 1 & 14.58 \(\pm\) 0.43&	35.49 \(\pm\) 0.75&	46.24 \(\pm\) 0.61&	56.97 \(\pm\) 0.38\\
\cmidrule{5-8}

 & \multirow{3}{*}{\rotatebox[origin=c]{0}{{\(\sim\) 2.3}}}& ResNet-26-16 & 1 & 13.02 \(\pm\) 0.89 &	34.71 \(\pm\) 0.97	& 47.61 \(\pm\) 0.71	& 59.07 \(\pm\) 0.38\\
& & ResNet-8-36 & 1 & 16.13 \(\pm\) 0.88	& 40.45 \(\pm\) 0.24	& 51.84 \(\pm\) 0.23	& 62.37 \(\pm\) 0.46\\
& & ResNet-8-16	& 5 & \textbf{17.7 \(\pm\) 0.08}	& \textbf{43.44 \(\pm\) 0.02}	& \textbf{54.78 \(\pm\) 0.23}	& \textbf{63.76 \(\pm\) 0.28}\\
\cmidrule{5-8}

&\multirow{3}{*}{\rotatebox[origin=c]{0}{{\(\sim\) 4.5}}} & ResNet-50-16 & 1 & 12.74 \(\pm\) 0.38	& 33.05 \(\pm\) 2.44	& 42.48 \(\pm\) 1.03	& 58.19 \(\pm\) 1.87\\
& & ResNet-8-50 & 1 & 16.53 \(\pm\) 0.78	& 41.36 \(\pm\) 1.2	& 53.55 \(\pm\) 0.19	& 64.37 \(\pm\) 0.41\\
& & ResNet-8-16 & 10 & \textbf{18.48 \(\pm\) 0.17}	& \textbf{45.15 \(\pm\) 0.11}	& \textbf{56.42 \(\pm\) 0.3}	& \textbf{64.98 \(\pm\) 0.01}\\
\cmidrule{5-8}

&\multirow{3}{*}{\rotatebox[origin=c]{0}{{\(\sim\) 9.5}}} & ResNet-110-16 & 1 & 12.53 \(\pm\) 0.24	& 29.68 \(\pm\) 0.77	& 39.88 \(\pm\) 0.85	& 53.08 \(\pm\) 1.57\\
& & ResNet-8-72 & 1   &16.51 \(\pm\)0.38	 &42.52 \(\pm\) 0.44	 &54.94 \(\pm\)0.8	 &\textbf{66.38 \(\pm\) 0.12} \\
& & ResNet-8-16 & 20 & \textbf{18.92 \(\pm\) 0.38}	&\textbf{46.56 \(\pm\) 0.41}	&\textbf{57.37 \(\pm\) 0.05}	&65.56 \(\pm\) 0.21 \\

\bottomrule
 
\end{tabular}}
\end{table}

\begin{table}[H]
    \centering
    \caption{Comparison between deep/wide CNNs and ensembles of less complex networks in terms of average sensitivity (\(||J(\mathbf{x})||_{F}\)) on the test set.
    \(M\) indicates the number of networks in the ensemble and \(N\) the number of training samples per class.
    These results are obtained using medium data augmentation (++). Means and standard deviations are obtained from five independent runs.
    We highlight the lowest values in bold.}
    \vskip 0.1in
    \label{tab:all_ds_sens}
    \resizebox{\textwidth}{!}
    {\scriptsize\begin{tabular}{cccccccc}
\toprule
\textbf{Dataset}& \textbf{Budget} \scriptsize{(MFLOPs)} &\textbf{Model} & \textbf{M} &\textbf{N} = 10 & \textbf{N} = 50  & \textbf{N} = 100 & \textbf{N} = 250 \\
\midrule

\multirow{6}{*}{\rotatebox[origin=c]{90}{{CIFAR-10}}}&\multirow{3}{*}{\rotatebox[origin=c]{0}{{\(\sim\) 9.5}}}& ResNet-110-16 & 1 & 0.42 \(\pm\) 0.03	&0.94 \(\pm\) 0.09	&1.02 \(\pm\) 0.15	&0.96 \(\pm\) 0.1 \\

& & ResNet-8-72 & 1 & 0.39 \(\pm\) 0.03	&0.74 \(\pm\) 0.03	&0.87 \(\pm\) 0.05	&1.03 \(\pm\) 0.02\\

& & ResNet-8-16 & 20 & \textbf{0.21 \(\pm\) 0.03}	&\textbf{0.51 \(\pm\) 0.03}	&\textbf{0.65 \(\pm\) 0.01}	&\textbf{0.77 \(\pm\) 0.04}\\

\cmidrule{5-8}

&\multirow{3}{*}{\rotatebox[origin=c]{0}{{\(\sim\) 4.5}}} & VGG-9-32	& 1 & 0.57 \(\pm\) 0.03	&0.81 \(\pm\) 0.06	&0.96 \(\pm\) 0.05	&1.18 \(\pm\) 0.03\\

& & VGG-5-76	& 1 & 0.36 \(\pm\) 0.03	&0.61 \(\pm\) 0.04	&0.69 \(\pm\) 0.02	&0.89 \(\pm\) 0.08\\

& & VGG-5-32	& 5 & \textbf{0.23 \(\pm\) 0.0}	&\textbf{0.4 \(\pm\) 0.03}	&\textbf{0.47 \(\pm\) 0.03}	&\textbf{0.6 \(\pm\) 0.02}\\

\midrule

\multirow{6}{*}{\rotatebox[origin=c]{90}{{CIFAR-100}}}&\multirow{3}{*}{\rotatebox[origin=c]{0}{{\(\sim\) 9.5}}}& ResNet-110-16 & 1 & 1.14 \(\pm\) 0.14	&1.8 \(\pm\) 0.24	&2.02 \(\pm\) 0.02	&2.45 \(\pm\) 0.16 \\
& & ResNet-8-72 & 1 & 1.2 \(\pm\) 0.04	&2.51 \(\pm\) 0.09	&3.08 \(\pm\) 0.02	&3.27 \(\pm\) 0.05\\
& & ResNet-8-16 & 20 & \textbf{0.54 \(\pm\) 0.02}	&\textbf{0.99 \(\pm\) 0.02}	&\textbf{1.16 \(\pm\) 0.0}	&\textbf{1.2 \(\pm\) 0.02}\\

\cmidrule{5-8}

&\multirow{3}{*}{\rotatebox[origin=c]{0}{{\(\sim\) 4.5}}} & VGG-9-32	& 1 & 1.29 \(\pm\) 0.16	&1.78 \(\pm\) 0.16	&2.28 \(\pm\) 0.79	&2.42 \(\pm\) 0.62\\
& & VGG-5-76	& 1 & 1.26 \(\pm\) 0.13	&2.26 \(\pm\) 0.09	&2.19 \(\pm\) 0.28	&2.47 \(\pm\) 0.24\\
& & VGG-5-32	& 5 & \textbf{0.76 \(\pm\) 0.03}	&\textbf{1.1 \(\pm\) 0.06}	&\textbf{1.39 \(\pm\) 0.17}	&\textbf{1.44 \(\pm\) 0.19}\\

\midrule

\multirow{3}{*}{\rotatebox[origin=c]{90}{SVHN}} & \multirow{3}{*}{\rotatebox[origin=c]{0}{{\(\sim\) 0.5}}}  & DenseNet-BC-52, k=12 & 1 & 0.15 \(\pm\) 0.1	&0.95 \(\pm\) 0.05	&0.79 \(\pm\) 0.1	&0.64 \(\pm\) 0.03\\
& & DenseNet-BC-16, k=30 & 1 & \textbf{0.11 \(\pm\) 0.07}	&1.09 \(\pm\) 0.12	&0.83 \(\pm\) 0.07	&0.65 \(\pm\) 0.05 \\
& & DenseNet-BC-16, k=12 & 6 & 0.14 \(\pm\) 0.06	&\textbf{0.83 \(\pm\) 0.08}	&\textbf{0.67 \(\pm\) 0.04}	&\textbf{0.54 \(\pm\) 0.03} \\

\bottomrule
 
\end{tabular}}
\end{table}